\documentclass{article}

\usepackage{PRIMEarxiv}

\usepackage[utf8]{inputenc} 
\usepackage[T1]{fontenc}    
\usepackage{hyperref}       
\usepackage{url}            
\usepackage{booktabs}       
\usepackage{amsfonts}       
\usepackage{nicefrac}       
\usepackage{microtype}      
\usepackage{lipsum}
\usepackage{fancyhdr}       
\usepackage{graphicx}       
\graphicspath{{media/}}     

\usepackage{adjustbox}
\usepackage{bbm}
\usepackage{bm}
\usepackage{amsmath}
\usepackage{booktabs}

\usepackage{makecell}
\usepackage{caption}
\usepackage{subcaption}
\usepackage{multirow}
\usepackage{xurl}
\usepackage[dvipsnames]{xcolor}

\usepackage{microtype}

\usepackage{array}
\usepackage{enumitem}
\usepackage{bm}
\usepackage{amsmath}
\usepackage{booktabs}
\usepackage{algorithm}
\usepackage{algpseudocode}
\usepackage{amssymb}
\usepackage{pifont}
\usepackage{tabularx}
\newcolumntype{Y}{>{\centering\arraybackslash}X}

\title{Word Embedding Dimension Reduction via Weakly-Supervised Feature Selection}

\usepackage{authblk}
\author[1]{Jintang Xue}
\author[1]{Yun-Cheng Wang}
\author[1]{Chengwei Wei}
\author[1]{C.-C. Jay Kuo}
\affil[1]{University of Southern California, Los Angeles, California, USA}

\begin{document}
\maketitle

\begin{abstract}
    As a fundamental task in natural language processing, word embedding converts each word into a representation in a vector space. A challenge with word embedding is that as the vocabulary grows, the vector space's dimension increases, which can lead to a vast model size. Storing and processing word vectors are resource-demanding, especially for mobile edge-devices applications. This paper explores word embedding dimension reduction. To balance computational costs and performance, we propose an efficient and effective weakly-supervised feature selection method named WordFS. It has two variants, each utilizing novel criteria for feature selection. Experiments on various tasks (e.g., word and sentence similarity and binary and multi-class classification) indicate that the proposed WordFS model outperforms other dimension reduction methods at lower computational costs. We have released the code for reproducibility along with the paper \footnote{https://github.com/jintangxue/WordFS}.
\end{abstract}

\section{Introduction}

In recent years, large language models (LLMs) have made a breakthrough in natural language processing (NLP). These models have revolutionized the understanding and generation of human language tasks such as question-answering, machine translation, and text summarization \cite{wu2023brief, SIP-2023-0064, naveed2023comprehensive}, etc. As a fundamental component in language models, word embedding represents words as vectors in a continuous, high-dimensional space \cite{SIP-2023-0010}. Word vectors capture semantic and syntactic meanings of words, providing word relationships for various downstream tasks. Static word embedding methods, such as Glove \cite{pennington2014glove}, Word2Vec \cite{mikolov2013efficient}, and Fasttext \cite{bojanowski2016enriching}, assign a fixed vector to each word. They convert input words to corresponding vectors for further processing and model training. Contextual word embedding methods, such as ELMo \cite{peters2018deep} and BERT \cite{devlin2018bert}, leverage deep-learning models to generate word vectors based on the context. A word can have a different contextual word embedding in a different sentence, allowing the model to capture subtle differences in the meaning of the same word in various contexts. 

A common challenge for static and contextual word embedding methods is that the high dimension of a word vector leads to an enormous model size. Typically, a word vector has hundreds to thousands of dimensions. For instance, storing a vocabulary of 3 million words in 300 dimensions would require 3.39 GB. Loading a 300-dimensional word embedding matrix with 2.5 million tokens would require up to 6 GB of memory on a 64-bit system \cite{raunak2019effective}. On the one hand, high-dimensional word vectors provide a good representation of complex human language, which is crucial for performing downstream tasks. On the other hand, high-dimensional word vectors have higher demands on computational resources and memory requirements. Thus, dimension reduction is critical in the application of word vectors.

Existing dimension reduction methods mainly consist of traditional PCA-based and deep-learning-based models. For PCA-based models, \cite{raunak2019effective} combines a post-processing technique with PCA to achieve an effective method for dimension reduction. For deep-learning-based models,  \cite{hwang2023embedtextnet} proposed a deep-learning method called EmbedTextNet for word embedding dimension reduction by leveraging a VAE model with a correlation penalty added to the weighted reconstruction loss. The model works well in low-dimensional embedding sizes but takes a long time to train. Other deep learning methods focus on model compression \cite{jiao2019tinybert, sanh2019distilbert} and quantization \cite{kim2020adaptive, shu2017compressing} rather than dimension reduction. 

Traditional unsupervised PCA-based methods are known for their efficiency and interpretability. In contrast, supervised deep-learning-based methods are inefficient and lack interpretability. For instance, autoencoders take longer training and inference time in sentence embedding dimension reduction than traditional methods \cite{zhang2024evaluating}. Semi-supervised feature selection methods can also be used for dimension reduction. Since word vectors can be viewed as extracted features for each word, feature selection methods have the potential to provide a straightforward yet effective way to reduce the word dimension. It could help balance unsupervised PCA-based methods and resource-intensive deep-learning-based methods.

This paper investigates using a small subset of labeled word similarity pairs to develop a weakly supervised dimension reduction method called WordFS (i.e., word dimension reduction with \textbf{F}eature \textbf{S}election). We demonstrate that one can achieve dimension reduction effectively by supervising a limited number of word similarity pairs. Note that word similarity is typically used in evaluation tasks but is rare in word embedding dimension reduction, a key novelty of this work. The proposed WordFS method consists of three stages: 1) post-processing, 2) feature extraction, and 3) weakly-supervised feature selection. We apply WordFS to other downstream tasks to show its generalizability. Experimental results show that WordFS outperforms existing methods in word similarity and various tasks while achieving much lower computational costs. 

This work has the following significant contributions.
    \begin{itemize}
    \item We propose a novel, effective, and efficient dimension reduction method called WordFS for word embeddings from the perspective of feature selection based on weakly-supervised learning.
    \item We demonstrate the potential of combining feature selection methods and word similarity for word embedding dimension reduction.
    \item We show the effectiveness and efficiency of our approach on various downstream tasks, including sentence similarity and classification tasks. Our method generally outperforms the existing techniques while being more straightforward and efficient.
    \end{itemize}
    
The rest of the paper is organized as follows. Related work is reviewed in Sec. \ref{sec:review}. The proposed WordFS method is described in Sec. \ref{sec:method}. Experimental results are shown in Sec. \ref{sec:experiments}. Finally, concluding remarks and future extensions are given in  Sec. \ref{sec:conclusion}.

\section{Related work} \label{sec:review}
Word embedding compression is an essential topic for storing and processing word embeddings, especially on computationally limited devices. Existing dimension reduction methods mainly consist of traditional and deep learning-based models.

\subsection{Matrix Decomposition Techniques}
Matrix decomposition techniques, such as singular value decomposition (SVD), principal components analysis (PCA),  non-negative matrix factorization (NMF), and factor analysis (FA), have been applied to dimension reduction of word embeddings. While simple and efficient, these methods are not highly effective.

Post-processing methods help enhance word embeddings in dimension reduction. Effective construction of lower dimensional word embeddings can be achieved by combining post-processing algorithms (PPAs) based on reducing the anisotropy \cite{mu2018allbutthetop} with the PCA method \cite{raunak2019effective}. The dimension-reduced vectors can perform similarly or even better than the original pre-trained word embeddings, outperforming most unsupervised methods. \cite{jha2021geodesic} extends this method to contextual embeddings by adding in additional geodesic distance information via the Isomap algorithm \cite{tenenbaum2000global}. PCA also performs well as an unsupervised dimension reduction method for pre-trained sentence embeddings \cite{zhang2024evaluating}.

\subsection{Non-Linear Dimensionality Reduction Techniques}
Non-linear dimensionality reduction techniques like uniform manifold approximation and projection (UMAP) and t-distributed stochastic neighbor embedding (t-SNE) have been used on dimension reduction tasks. UMAP is a dimensionality reduction method based on manifold theory and topological data analysis. It constructs a topological representation of high-dimensional data by approximating local manifolds and stitching together their fuzzy simplicial sets. However, UMAP's performance in reducing the dimensionality of word embeddings is not satisfactory. t-SNE is a statistical method based on stochastic neighbor embedding. It assigns each data point a location in a two- or three-dimensional map for visualizing high-dimensional data. Applying t-SNE to a large corpus while keeping many dimensions requires a significant amount of time and memory, making it unsuitable for this task.

\subsection{Deep Learning Methods}
Deep learning-based models have recently become quite popular, and efforts have been made to explore their potential in reducing word embedding dimensions. EmbedTextNet \cite{hwang2023embedtextnet} achieves better performance in low-dimensional embedding sizes by utilizing a VAE model with a correlation penalty added to the weighted reconstruction loss. However, deep learning models typically require more computational resources and longer training time, contradicting the original goal of dimension reduction.

\subsection{Feature Selection Methods}
Feature selection is another simple and intuitive method for removing redundant features. Feature selection methods can be applied to bag-of-words representations, where each unique word is treated as a distinct feature, for emotion recognition \cite{erenel2020new} and sentiment classification \cite{uysal2017sentiment} tasks to reduce the number of terms and save storage and memory space. Also, feature selection methods can be applied for text classification to select the most representative word \cite{rui2016unsupervised}. However, the role of feature selection for word embedding dimension reduction is under-explored. This paper shows that a straightforward and intuitive feature selection method can effectively reduce dimension. We believe applying the feature selection method for word embedding dimension reduction is novel.

\section{Proposed Method} \label{sec:method}
\subsection{System Overview}
\begin{figure}[t]
\centering
\includegraphics[width=0.4\textwidth]{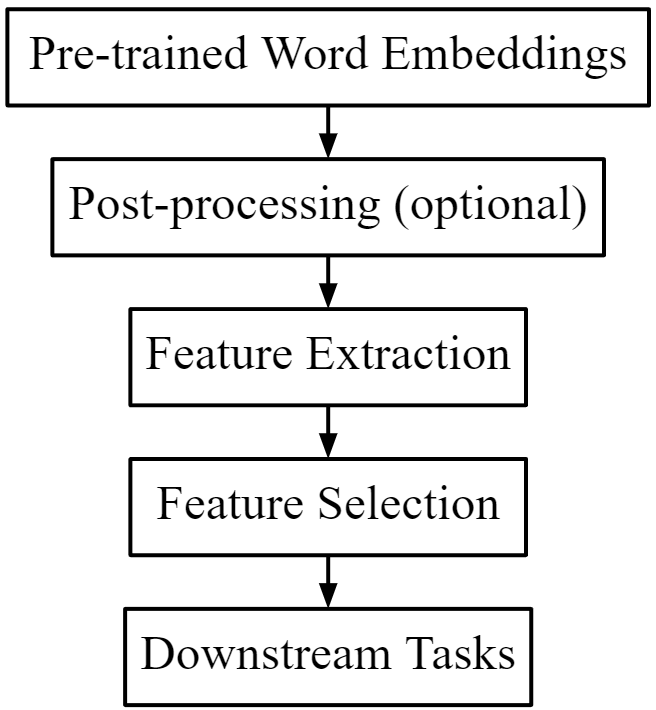}
\caption{An overview of the proposed WordFS method.}\label{fig:pipeline}
\end{figure}

Our WordFS method consists of three stages: 1) post-processing, 2) feature extraction, and 3) feature selection, as shown in Figure \ref{fig:pipeline}. We adopt word similarity datasets as the source of supervision to guide the feature selection in our model. As a widely-used benchmark of word embeddings, the word similarity task evaluates the quality of word embeddings in representing semantic and syntactic meanings by measuring how closely the representation of word vectors matches human perception of similarity \cite{wang2019evaluating}, making them good supervision resources for word embedding dimension reduction. Also, since word similarity captures fundamental characteristics of word embeddings, which are crucial for many NLP tasks, its potential for generalizing to downstream tasks is noteworthy.

In our method, We begin by optionally applying a post-processing method to our pre-trained word embeddings since the anisotropy may not always be harmful \cite{ait2023anisotropy}. Word vectors pre-trained on a smaller amount of data may be susceptible to noise and could benefit from post-processing. On the other hand, word vectors pre-trained on a larger corpus of data may capture important nuanced information necessary for downstream tasks but could be negatively impacted by post-processing. We then extract pair-wise features for each word pair by conducting element-wise production with normalization. 
Then, we adopt two different criteria for feature selection to evaluate each dimension and identify essential dimensions. The first one is based on a supervised feature selection method called RFT \cite{yang2022supervised}. We select mean squared error (MSE) as our cost function. The second is based on Spearman's rank correlation coefficient between each feature dimension and word similarity scores.
Finally, the selected dimensions of the word embeddings are kept, and other dimensions are discarded according to the requirements for dimension reduction. The details of each stage are elaborated below.

\subsection{Post-processing}
To improve the quality of word embedding in downstream applications, performing post-processing techniques on pre-trained word embeddings can be crucial. The post-processing algorithm (PPA) \cite{mu2018allbutthetop} is an effective post-processing method to improve the isotropy of word representations by removing the common mean vector and the top principal components of all words. Details of PPA are shown in Algorithm ~\ref{alg:ppa} \cite{mu2018allbutthetop}. First, the mean vector is computed from all embeddings in the vocabulary and then subtracted from each embedding. Next, the PCA components are derived from the adjusted mean matrix, with the top-D components being chosen. Finally, the projections on the top-D principal components are removed from each word embedding.

The existing effective PCA-based method \cite{raunak2019effective} applies the PPA before and after dimension reduction to enhance both the original and the dimension-reduced word vectors. In our method, we only apply PPA before dimension reduction, which makes our method simpler. Applying PCA-based post-processing to the selected subset after feature selection may disrupt the previous results because PCA and the feature selection criteria can have different objectives. Specifically, PCA transforms features to maximize variance, while feature selection methods usually select features based on their correlation with labels. Also, the selected feature subset may lose crucial information for PCA to find valuable principal components. As an enhancement of original word embeddings before dimension reduction, we also make PPA an optional choice. The reason is that anisotropy may not always be harmful \cite{ait2023anisotropy}, and using PPA depends on the pre-trained word embeddings and specific application scenarios. Specifically, word embeddings trained on less data may be vulnerable to noise. Applying PPA helps mitigate some of the noise and offers better performance. On the other hand, using PPA may result in the loss of crucial and intricate information related to word representation, which can harm performance. Word embeddings trained on relatively small data benefit from uniformly distributed word embeddings, while complex contextual tasks may require anisotropy to capture detailed information.

\begin{algorithm}
\caption{Post-processing Algorithm (PPA)}
\label{alg:ppa}
\begin{algorithmic}[1]
\State \textbf{Input:} Word vector matrix $M$; Threshold $D$
\State Subtract the mean: 
\Statex $n \gets$ number of word vectors in $M$ 
\Statex $\mu \gets \frac{1}{n} \sum_{i=1}^{n} M_i$
\Statex \textbf{for each word vector} $M_i$ \textbf{in} $M$ \textbf{do}
\Statex \hspace{\algorithmicindent} $M_i \gets M_i - \mu$
\Statex \textbf{end}
\State Compute the PCA components: 
\Statex $U \gets PCA(M)$ 
\Statex \textbf{for} $i = 1$ \textbf{to} $D$ \textbf{do}
\Statex \hspace{\algorithmicindent} $u_i \gets$ the $i^{th}$ principal component from $U$
\Statex \textbf{end}
\State Remove the top-$D$ principal components:
\Statex \textbf{for all} {$v$ in $M$} \textbf{do}
\Statex \hspace{\algorithmicindent} $v' \gets v - \sum_{i=1}^{D} (u_i^\top v)u_i$
\Statex \textbf{end}
\State \textbf{Output:} Enhanced word vector matrix $M'$
\end{algorithmic}
\end{algorithm}

\subsection{Feature Extraction}
In this stage, we extract suitable features for feature selection based on the dataset that provides supervision.
In this paper, we leverage word similarity data as the weak supervision of our model. Since the word similarity is based on the correlation (i.e., word similarity score) between every two words, we extract pair-wise features for each word similarity pair. 

The pair-wise features are constructed in each dimension based on cosine similarity, an effective measure for evaluating the similarity between word embeddings. Cosine similarity between two-word vectors is derived by dividing their dot product by the product of their magnitudes:

\begin{equation}
\begin{aligned}
\text{Sim}(\mathbf{a}, \mathbf{b}) &= \frac{\mathbf{a} \cdot \mathbf{b}}{\|\mathbf{a}\|\|\mathbf{b}\|} \\
&= \frac{\sum_{i=1}^{n} a_i b_i}{\sqrt{\sum_{i=1}^{n} a_i^2} \cdot \sqrt{\sum_{i=1}^{n} b_i^2}},
\end{aligned}
\end{equation}

where $\mathbf{a}$ and $\mathbf{b}$ are embeddings of the two words, and n represents the number of elements in a word embedding. $a_i$ and $b_i$ are the $i$th elements of $\mathbf{a}$ and $\mathbf{b}$, respectively. The cosine similarity score can be viewed as the sum of the normalized results of element-wise products. Each pair of elements (i.e., each dimension) has a linear impact on the cosine similarity score of that pair of word vectors. Therefore, the normalized results of element-wise products can be viewed as features to predict the similarity score. The feature we use for each dimension is:

\begin{equation}
\begin{aligned}
f_i &= \frac{a_i b_i}{\sqrt{\sum_{i=1}^{n} a_i^2} \cdot \sqrt{\sum_{i=1}^{n} b_i^2}},
\end{aligned}
\end{equation}

where $f_i$ is the feature of the $i^{th}$ dimension extracted from thecorresponding elements $a_i$ and $b_i$ of the two word embeddings.

\subsection{Feature Selection}
The method for the feature selection process depends on the objective of downstream tasks. Specifically, we employ two feature selection methods for prediction and similarity tasks. These two feature selection methods are developed into our model's variants, WordFS-S and WordFS-P. We focus on filter feature selection techniques since they are simple and computationally efficient \cite{sheikhpour2017survey}. Although different dimensions of distributed word embeddings may capture nuanced and interdependent semantic relationships, they allow us to identify dimensions that contribute most significantly to model performance. Besides, filter feature selection techniques select features based on their correlation with the labels and are independent of the classifier. As a result, the selected features are generalizable.

\textbf{Prediction tasks:}
 For prediction tasks, we leverage the methods from Discriminant Feature Test (DFT) and Relevant Feature Test (RFT) \cite{yang2022supervised} to build our WordFS-P model. DFT and RFT are a pair of filter feature selection techniques proposed recently for classification and regression tasks, respectively. They have been widely used in green learning architectures \cite{kuo2023green} for feature dimension reduction to reach a smaller model size. In our experiments, RFT is utilized for feature selection, and we adopt the word similarity scores as the labels. RFT partitions each feature dimension into two subintervals and calculates the overall mean square error (MSE) as the loss function. A smaller loss function means a better feature dimension. Specifically, given a feature of the $i^{th}$ dimension $f_i$, the feature space is partitioned into $B=2^k, k=1,2,...$ uniform segments and the optimal partition threshold is searched among $B-1$ candidates in the range $[min(f_i), max(f_i)]$: 

\begin{equation}
\begin{aligned}
f_{b}^{i} &= min(f_i) + \frac{b}{B} \left[ max(f_i) - min(f_i) \right],
\end{aligned}
\end{equation}

where $b=1,2,...B-1$. A threshold $t$ partitions the $i^{th}$ feature space into the left subset $S_{L,t}^{i}$ and the right subset $S_{R,t}^{i}$. The MSEs for regression $R_{L,t}^{i}$ and $R_{R,t}^{i}$ are separately calculated in the two subsets by using the mean of target values as the estimated regression value of all samples. Then, the RFT loss is defined as the weighted sum of the two MSEs:

\begin{equation}
R_{t}^{i} = \frac{N_{L,t}^{i} R_{L,t}^{i} + N_{R,t}^{i} R_{R,t}^{i}}{N},
\end{equation}

where $N_{L,t}^{i}$ and $N_{R,t}^{i}$ denote the number of samples in each subset and $N$ is the total number of samples. The RFT loss of the $i$th feature dimension is defined as the minimal RFT loss among all the candidate thresholds:
\begin{equation}
R_i = \min_{t \in T} R_{t}^{i}.
\end{equation}
Finally, all feature dimensions' RFT loss is ranked in ascending order, and the top $K$ dimensions are selected, where $K$ is the dimension of the word vectors after our dimension reduction approach.

\textbf{Similarity tasks:} 
 Similarity tasks are usually done by calculating the cosine similarity between target vectors, which differs from prediction tasks. The cost function of DFT/RFT may not match well with the evaluation criteria of similarity tasks. Inspired by the evaluation of word similarity, we use Spearman's rank correlation coefficient to develop our WordFS-S model. It involves finding the correlation between the ranks generated by features extracted from each dimension and the target labels. First, the ranked values of each feature dimension and the labels are calculated. Second, Spearman's rank correlation coefficient for each feature dimension is calculated by the Pearson correlation coefficient between the ranked values of each dimension and the labels:
\begin{equation}
r_{i} = \rho_{R(f_i),R(y)} = \frac{\text{cov}(R(f_i), R(y))}{\sigma_{R(f_i)} \sigma_{R(y)}},
\end{equation}
where $R(\cdot)$ represents the rank of given raw scores. The higher the correlation coefficient, the better the feature dimension is. Finally, the Spearman's rank correlation coefficients of all the feature dimensions are then ranked in descending order, and the top $K$ dimensions are selected.

\section{Experiments} \label{sec:experiments}
We evaluate our method by applying it to several pre-trained word embeddings. The word similarity datasets provide weak supervision to guide our feature selection module. The dimension-reduced word embeddings are used for word similarity tasks and downstream tasks. Then, we compare the results of our method with the original pre-trained word embeddings, the reduced word embeddings from UMAP \cite{mcinnes2018umap}, the dimension-reduced word embeddings from the PCA-based method called Algo \cite{raunak2019effective}, and the results from the deep learning-based method called EmbedTextNet \cite{hwang2023embedtextnet}. The performance of different datasets can vary due to their specific requirements and emphasis. Finding a general dimensionality reduction method that performs well across all datasets without fine-tuning is challenging. Therefore, we focus on comparing the average performance for each task. The method that achieves better performance on average is considered better.

\subsection{Pre-trained Word Embeddings}
In our experiments, we use the following three pre-trained word embeddings. 
\begin{enumerate}
\item Glove word embeddings \cite{pennington2014glove} trained on Wikipedia 2014 and Gigaword 5 corpus (6B tokens, 400K vocabulary) \\
They are available in multiple dimensions, precisely 50, 100, 200, and 300. 
\item Word2vec word embeddings \cite{mikolov2013efficient} trained on a portion of the Google News dataset (about 100B words) \\
The model provides word vectors of 300 dimensions for 3M words and phrases. 
\item Fasttext word embeddings \cite{bojanowski2016enriching} trained on Wikipedia 2017, UMBC webbase corpus, and statmt.org news dataset (16B tokens) \cite{mikolov2018advances} \\
It generates 1 million 300-dimensional word vectors. The Gloved-based model is trained on the least amount of data among the three pre-trained word embeddings.
\end{enumerate}

\subsection{Word Similarity Datasets}
We use the widely-used word similarity datasets \cite{faruqui2014improving} to evaluate our method. The word similarity datasets contain word pairs and their corresponding scores from human annotators based on perceived relatedness or similarity. The cosine similarity of each pair of words calculates the similarity score from the word embeddings. We use Spearman's rank correlation coefficient as our evaluation metric. It measures how closely the ranking derived from the cosine similarities of given word vectors matches those based on human judgments. A higher value of this metric indicates a better match to human-labeled similarity rankings.

We first evaluate our methods on twelve word similarity datasets based on all the three word embeddings. In later experiments, we will assess the generalizability of our WordFS model across various embeddings on multiple downstream tasks.
Since we choose word similarity as the guidance for feature selection, we apply 5-fold cross-validation to each word similarity dataset to train and test our method. The reported results are the average Spearman's rank correlation coefficients from the five folds. We evaluate the pre-trained word embeddings and all the methods on the same cross-validation fold and take the average for a fair comparison. All the results we report average over five different cross-validation trials in all the experiments. We set the number of bins in RFT to 4, the default and recommended value from the RFT model, and the threshold in the PPA to 7, the same as in the PCA-based model.

\begin{table}
\begin{center}
\caption{\label{all_word_similarity_glove}
Comparison of Spearman's rank correlation coefficient across multiple word similarity datasets, where \textbf{Boldface} indicates the best value in each column, and \underline{underline} indicates the second best value in each column.
}
\setlength{\tabcolsep}{7pt} 
\renewcommand{\arraystretch}{1.2} 
\resizebox*{\linewidth}{!}{
\begin{tabular}{c|c|c|c|c|c|c|c|c|c|c|c|c|c}
\noalign{\hrule height 2pt}
 & \textbf{MC} & \textbf{Men} & \textbf{M} & \textbf{M} & \textbf{RG} & \textbf{RW} & \textbf{Sim} & \textbf{VE} & \textbf{WS} & \textbf{WS} & \textbf{WS} & \textbf{YP} &\\
 \textbf{Dataset} & \textbf{-30} & \textbf{-TR} & \textbf{Turk} & \textbf{Turk} & \textbf{-65} & \textbf{Stan} & \textbf{Lex} & \textbf{RB} & \textbf{-353} & \textbf{-353} & \textbf{-353} & \textbf{-130} & \textbf{Avg}\\
 &  & \textbf{-3k} & \textbf{-287} & \textbf{-771} &  & \textbf{ford} & \textbf{-999} & \textbf{-143} & \textbf{-ALL} & \textbf{-REL} & \textbf{-SIM} & &\\
\noalign{\hrule height 2pt}
\text{Glove-300D} & \text{64.03} & \text{73.71} & \text{62.44} & \text{64.60} & \text{72.97} & \text{41.16} & \text{36.97} & \text{29.97} & \text{60.27} & \text{55.44} & \text{64.74} & \text{55.15} & \text{56.79}\\
\hline
\text{UMAP-150D} & \text{25.39} & \text{38.81} & \text{35.88} & \text{31.54} & \text{32.50} & \text{13.91} & \text{25.33} & \text{24.13} & \text{30.76} & \text{20.19} & \text{38.23} & \text{14.00} & \text{27.56}\\
\text{Algo-150D} & \text{67.97} & \text{75.21} & \text{62.47} & \text{64.11} & \underline{72.94} & \text{43.45} & \text{37.81} & \textbf{38.06} & \text{66.71} & \underline{60.55} & \underline{70.74} & \textbf{58.09} & \text{59.84}\\
\text{EmbedTextNet-150D} & \textbf{79.59} & \text{75.43} & \text{58.36} & \text{62.91} & \textbf{80.00} & \text{41.31} & \text{37.49} & \text{27.33} & \text{63.06} & \text{55.35} & \text{67.50} & \text{55.80} & \text{58.68}\\
\text{WordFS-P-woP-150D(ours)} & \text{66.79} & \text{74.94} & \text{60.68} & \text{63.44} & \text{69.83} & \text{36.16} & \text{37.93} & \text{30.26} & \text{62.49} & \text{53.69} & \text{66.78} & \text{52.27} & \text{56.27}\\
\text{WordFS-P-wP-150D(ours)} & \underline{73.63} & \text{75.38} & \underline{64.38} & \underline{64.24} & \text{70.37} & \underline{45.01} & \text{42.47} & \text{34.88} & \textbf{68.53} & \textbf{60.58} & \text{70.25} & \text{50.54} & \underline{60.02}\\
\text{WordFS-S-woP-150D(ours)} & \text{67.79} & \underline{75.78} & \text{62.62} & \text{62.95} & \text{70.26} & \text{36.07} & \underline{44.72} & \underline{36.61} & \text{64.72} & \text{55.36} & \text{69.42} & \text{45.76} & \text{57.67}\\
\text{WordFS-S-wP-150D(ours)} & \text{72.27} & \textbf{75.96} & \textbf{65.57} & \textbf{64.35} & \text{70.69} & \textbf{45.22} & \textbf{45.08} & \text{36.03} & \underline{67.16} & \text{59.18} & \textbf{71.00} & \underline{52.90} & \textbf{60.45}\\
\hline
\text{Glove-100D} & \text{57.18} & \text{68.08} & \text{60.78} & \text{57.81} & \text{65.29} & \text{36.54} & \text{29.79} & \text{30.61} & \text{52.83} & \text{47.62} & \text{58.55} & \text{43.02} & \text{50.68}\\
\text{UMAP-100D} & \text{22.88} & \text{39.72} & \text{38.15} & \text{31.06} & \text{40.10} & \text{14.97} & \text{26.08} & \text{18.97} & \text{30.78} & \text{21.48} & \text{37.27} & \text{14.98} & \text{28.04}\\
\text{Algo-100D} & \text{69.99} & \text{70.64} & \text{60.65} & \textbf{60.92} & \textbf{75.93} & \text{39.55} & \text{35.80} & \text{29.70} & \text{62.33} & \text{53.29} & \text{68.49} & \textbf{50.59} & \text{56.49}\\
\text{EmbedTextNet-100D} & \underline{70.21} & \text{70.43} & \underline{62.04} & \text{59.28} & \underline{74.42} & \text{37.20} & \text{35.19} & \text{30.63} & \text{59.66} & \text{54.10} & \text{64.94} & \text{44.79} & \text{55.24}\\
\text{WordFS-P-wP-100D(ours)} & \textbf{72.91} & \underline{73.96} & \text{61.16} & \underline{60.54} & \text{69.91} & \textbf{43.65} & \underline{42.65} & \underline{34.93} & \textbf{67.80} & \textbf{58.91} & \underline{69.15} & \underline{46.28} & \textbf{58.49}\\
\text{WordFS-S-wP-100D(ours)} & \text{69.63} & \textbf{74.72} & \textbf{65.29} & \text{60.23} & \text{68.79} & \underline{43.16} & \textbf{43.95} & \textbf{37.17} & \underline{66.11} & \underline{56.79} & \textbf{69.65} & \text{46.06} & \underline{58.46}\\
\hline
\text{Glove-50D} & \text{53.75} & \text{65.24} & \textbf{60.67} & \textbf{55.14} & \text{58.41} & \text{34.02} & \text{26.44} & \text{25.26} & \text{49.82} & \text{44.94} & \text{55.66} & \text{36.03} & \text{47.11}\\
\text{UMAP-50D} & \text{17.39} & \text{39.84} & \text{36.15} & \text{31.14} & \text{41.56} & \text{15.13} & \text{26.22} & \text{22.71} & \text{31.37} & \text{20.87} & \text{38.87} & \text{8.59} & \text{27.49}\\
\text{Algo-50D} & \underline{64.48} & \text{61.43} & \text{53.19} & \text{47.30} & \underline{63.65} & \text{33.65} & \text{26.75} & \textbf{34.77} & \text{55.60} & \text{45.01} & \text{59.35} & \textbf{38.00} & \text{48.60}\\
\text{EmbedTextNet-50D} & \text{55.58} & \text{65.90} & \text{63.61} & \text{55.18} & \text{53.97} & \text{32.60} & \text{28.57} & \underline{32.21} & \text{57.51} & \text{50.19} & \text{64.01} & \text{36.18} & \text{49.63}\\
\text{WordFS-P-wP-50D(ours)} & \text{63.11} & \underline{69.91} & \text{55.14} & \underline{52.58} & \text{63.52} & \textbf{39.00} & \textbf{41.33} & \text{30.49} & \textbf{64.35} & \underline{53.76} & \underline{64.28} & \underline{37.64} & \underline{52.92}\\
\text{WordFS-S-wP-50D(ours)} & \textbf{67.81} & \textbf{71.06} & \underline{58.97} & \text{52.08} & \textbf{64.96} & \underline{36.62} & \underline{39.96} & \text{30.28} & \underline{60.46} & \textbf{56.47} & \textbf{66.33} & \text{34.86} & \textbf{53.32}\\
\noalign{\hrule height 2pt}
\text{Word2vec-300D} & \text{73.42} & \text{77.01} & \text{68.52} & \text{66.73} & \text{71.65} & \text{53.39} & \text{44.19} & \text{49.69} & \text{68.92} & \text{61.42} & \text{76.87} & \text{53.80} & \text{63.80}\\
\hline
\text{UMAP-150D} & \text{36.38} & \text{50.43} & \text{42.53} & \text{26.17} & \text{35.13} & \text{26.00} & \text{22.32} & \text{6.25} & \text{42.09} & \text{32.59} & \text{51.84} & \text{17.37} & \text{32.42}\\
\text{Algo-150D} & \textbf{78.91} & \textbf{78.23} & \underline{63.46} & \textbf{65.86} & \textbf{74.93} & \text{51.43} & \text{42.53} & \text{41.13} & \text{68.25} & \underline{60.28} & \textbf{75.11} & \text{48.34} & \text{62.37}\\
\text{EmbedTextNet-150D} & \text{72.28} & \text{76.33} & \text{63.09} & \underline{65.28} & \underline{72.33} & \underline{52.27} & \text{43.53} & \text{44.47} & \text{66.44} & \text{57.95} & \underline{75.04} & \text{48.89} & \text{61.49}\\
\text{WordFS-P-woP-150D(ours)} & \text{68.16} & \text{76.16} & \text{63.35} & \text{63.39} & \text{70.79} & \text{51.60} & \text{45.32} & \textbf{55.99} & \text{67.68} & \text{58.04} & \text{72.44} & \text{48.14} & \text{61.75}\\
\text{WordFS-P-wP-150D(ours)} & \text{66.80} & \underline{76.77} & \text{63.43} & \text{63.45} & \text{69.99} & \text{52.00} & \text{46.57} & \text{50.77} & \text{66.81} & \text{56.57} & \text{72.64} & \text{48.75} & \text{61.21}\\
\text{WordFS-S-woP-150D(ours)} & \underline{75.71} & \text{75.93} & \textbf{64.22} & \text{63.35} & \text{71.63} & \text{51.86} & \textbf{46.95} & \underline{51.89} & \textbf{69.58} & \textbf{61.30} & \text{72.85} & \textbf{53.07} & \textbf{63.19}\\
\text{WordFS-S-wP-150D(ours)} & \text{74.80} & \text{76.73} & \text{61.72} & \text{63.74} & \text{70.55} & \textbf{52.59} & \underline{46.77} & \text{49.56} & \underline{68.79} & \text{58.41} & \text{74.73} & \underline{51.99} & \underline{62.53}\\
\hline
\text{UMAP-100D} & \text{33.18} & \text{51.38} & \text{42.27} & \text{27.97} & \text{33.24} & \text{26.86} & \text{23.03} & \text{8.13} & \text{42.56} & \text{31.03} & \text{53.53} & \text{11.16} & \text{32.03}\\
\text{Algo-100D} & \textbf{80.06} & \textbf{76.08} & \textbf{63.86} & \textbf{63.56} & \textbf{72.26} & \text{48.21} & \text{39.76} & \text{30.78} & \underline{67.49} & \textbf{59.81} & \textbf{73.40} & \text{39.65} & \text{59.58}\\
\text{EmbedTextNet-100D} & \underline{74.11} & \underline{75.73} & \underline{62.21} & \underline{63.14} & \underline{71.10} & \textbf{50.33} & \text{41.89} & \text{40.12} & \text{66.10} & \underline{59.38} & \underline{72.89} & \underline{45.34} & \underline{60.19}\\
\text{WordFS-P-woP-100D(ours)} & \text{62.45} & \text{74.14} & \text{57.35} & \text{61.32} & \text{69.55} & \underline{50.01} & \underline{45.18} & \underline{52.28} & \text{64.82} & \text{55.96} & \text{71.10} & \text{44.29} & \text{59.04}\\
\text{WordFS-S-woP-100D(ours)} & \text{65.42} & \text{74.45} & \text{60.81} & \text{60.28} & \text{66.60} & \text{49.75} & \textbf{45.70} & \textbf{52.60} & \textbf{68.49} & \text{57.76} & \text{70.23} & \textbf{50.48} & \textbf{60.21}\\
\hline
\text{UMAP-50D} & \text{28.38} & \text{50.06} & \text{42.88} & \text{27.44} & \text{35.31} & \text{25.90} & \text{22.21} & \text{9.78} & \text{44.41} & \text{34.64} & \text{54.23} & \text{16.77} & \text{32.67}\\
\text{Algo-50D} & \textbf{73.87} & \text{70.36} & \underline{62.29} & \text{54.36} & \textbf{66.97} & \text{41.89} & \text{32.36} & \text{22.72} & \text{59.75} & \text{53.79} & \underline{66.67} & \underline{37.28} & \underline{53.53}\\
\text{EmbedTextNet-50D} & \underline{72.49} & \textbf{73.10} & \textbf{66.33} & \textbf{59.24} & \text{62.16} & \textbf{46.30} & \text{36.71} & \text{26.65} & \textbf{63.99} & \textbf{55.74} & \textbf{72.62} & \text{35.26} & \textbf{55.88}\\
\text{WordFS-P-woP-50D(ours)} & \text{49.65} & \text{69.72} & \text{53.24} & \underline{55.01} & \underline{63.96} & \text{44.47} & \underline{43.12} & \underline{41.79} & \text{57.92} & \text{47.89} & \text{65.01} & \text{34.51} & \text{52.19}\\
\text{WordFS-S-woP-50D(ours)} & \text{54.68} & \underline{70.87} & \text{51.66} & \text{54.62} & \text{55.86} & \underline{44.65} & \textbf{43.52} & \textbf{47.26} & \underline{61.96} & \underline{49.05} & \text{65.16} & \textbf{38.93} & \text{53.18}\\
\noalign{\hrule height 2pt}
\text{Fasttext-300D} & \text{77.08} & \text{79.01} & \text{69.84} & \text{70.57} & \text{82.68} & \text{52.23} & \text{45.11} & \text{43.86} & \text{70.63} & \text{64.65} & \text{79.66} & \text{48.92} & \text{65.35}\\
\hline
\text{UMAP-150D} & \text{56.94} & \text{59.32} & \text{46.98} & \text{43.50} & \text{49.67} & \text{33.78} & \text{29.16} & \text{32.07} & \text{40.20} & \text{29.54} & \text{54.10} & \text{20.65} & \text{41.33}\\
\text{Algo-150D} & \textbf{92.63} & \underline{80.59} & \textbf{71.32} & \textbf{70.74} & \textbf{87.97} & \text{51.11} & \text{45.04} & \underline{45.03} & \text{73.18} & \underline{68.03} & \text{77.21} & \text{49.31} & \textbf{67.68}\\
\text{EmbedTextNet-150D} & \underline{77.54} & \text{78.50} & \underline{68.39} & \text{69.78} & \underline{82.10} & \text{52.03} & \text{45.71} & \text{38.11} & \text{68.01} & \text{62.98} & \text{74.77} & \text{51.57} & \text{64.12}\\
\text{WordFS-P-woP-150D(ours)} & \text{70.45} & \text{79.52} & \text{65.50} & \text{67.05} & \text{80.15} & \text{51.29} & \text{46.59} & \text{39.11} & \text{69.56} & \text{60.91} & \text{77.27} & \text{51.88} & \text{63.27}\\
\text{WordFS-P-wP-150D(ours)} & \text{76.17} & \text{80.13} & \text{67.26} & \underline{70.29} & \text{79.83} & \textbf{54.58} & \underline{47.44} & \text{38.84} & \text{72.93} & \text{66.76} & \textbf{78.53} & \textbf{55.68} & \text{65.70}\\
\text{WordFS-S-woP-150D(ours)} & \text{69.54} & \text{80.39} & \text{64.20} & \text{69.73} & \text{75.89} & \text{53.47} & \text{47.43} & \text{44.62} & \underline{74.32} & \text{66.94} & \text{77.07} & \text{52.69} & \text{64.69}\\
\text{WordFS-S-wP-150D(ours)} & \text{73.42} & \textbf{80.89} & \text{66.50} & \text{68.87} & \text{78.78} & \underline{53.94} & \textbf{49.14} & \textbf{46.14} & \textbf{74.99} & \textbf{71.09} & \underline{77.40} & \underline{53.71} & \underline{66.24}\\
\hline
\text{UMAP-100D} & \text{55.58} & \text{59.31} & \text{46.16} & \text{43.69} & \text{41.73} & \text{33.21} & \text{28.80} & \text{32.09} & \text{39.17} & \text{29.49} & \text{53.00} & \text{19.40} & \text{41.33}\\
\text{Algo-100D} & \textbf{89.65} & \textbf{78.80} & \textbf{71.86} & \underline{66.88} & \textbf{87.66} & \text{49.19} & \text{42.30} & \text{39.96} & \text{70.33} & \text{64.11} & \text{74.89} & \text{45.88} & \textbf{65.13}\\
\text{EmbedTextNet-100D} & \underline{81.65} & \text{77.74} & \underline{68.26} & \textbf{66.93} & \underline{81.72} & \underline{51.34} & \text{40.62} & \underline{41.30} & \text{69.21} & \text{62.20} & \underline{75.68} & \text{46.99} & \text{63.64}\\
\text{WordFS-P-wP-100D(ours)} & \text{70.91} & \text{77.40} & \text{63.64} & \text{66.07} & \text{76.81} & \textbf{52.08} & \underline{46.48} & \text{35.23} & \underline{71.05} & \underline{65.77} & \text{74.08} & \underline{52.72} & \text{62.69}\\
\text{WordFS-S-wP-100D(ours)} & \text{75.71} & \underline{78.33} & \text{65.38} & \text{66.44} & \text{76.89} & \text{51.25} & \textbf{47.24} & \textbf{43.04} & \textbf{72.63} & \textbf{67.27} & \textbf{75.96} & \textbf{56.67} & \underline{64.73}\\
\hline
\text{UMAP-50D} & \text{55.12} & \text{58.52} & \text{46.70} & \text{43.67} & \text{40.37} & \text{33.63} & \text{29.38} & \text{31.36} & \text{40.87} & \text{31.62} & \text{54.07} & \text{21.06} & \text{40.53}\\
\text{Algo-50D} & \textbf{79.37} & \textbf{72.74} & \textbf{70.66} & \textbf{61.06} & \textbf{77.50} & \text{45.47} & \text{35.79} & \text{37.48} & \text{59.66} & \text{52.72} & \text{67.79} & \text{43.79} & \text{58.67}\\
\text{EmbedTextNet-50D} & \underline{74.80} & \underline{72.39} & \underline{69.21} & \underline{60.69} & \underline{73.40} & \text{45.50} & \text{36.04} & \textbf{42.57} & \textbf{70.13} & \textbf{62.00} & \underline{76.17} & \text{34.52} & \textbf{59.78}\\
\text{WordFS-P-wP-50D(ours)} & \text{61.31} & \text{70.80} & \text{54.88} & \text{58.09} & \text{67.07} & \textbf{46.60} & \underline{44.66} & \text{26.71} & \text{64.36} & \underline{57.75} & \text{66.41} & \underline{50.27} & \text{55.74}\\
\text{WordFS-S-wP-50D(ours)} & \text{67.94} & \text{71.97} & \text{62.42} & \text{60.33} & \text{67.92} & \underline{45.53} & \textbf{45.48} & \underline{41.89} & \underline{67.41} & \text{59.24} & \textbf{70.99} & \textbf{54.36} & \underline{59.62}\\
\noalign{\hrule height 2pt}
\end{tabular}
}
\end{center}
\end{table}

Table~\ref{all_word_similarity_glove} shows the results across twelve word similarity datasets of different dimension reduction methods reduced from 300-dimensional Glove word embeddings to 150, 100, and 50 dimensions, respectively. In the table, WordFS-P and WordFS-S represent our proposed methods based on RFT and similarity feature selection methods, respectively. P represents PPA, and wP and woP represent PPA and without PPA. Algo is the method proposed in \cite{raunak2019effective}, which utilizes PPA before and after PCA. The last column is the average of the previous columns. The UMAP results are significantly lower compared to other methods. This is understandable because UMAP is a general dimension reduction method and is not specifically designed to reduce the dimension of word embeddings, unlike other methods. For the dimension-reduced vectors with dimensions of 50, 100, and 150 based on the Glove embedding, our methods outperform the original Glove word embeddings and other methods in most datasets across these reduced dimensions. Our WordFS-S-wP method achieves the highest average score when the dimension is reduced to 50 and 150 and is close to the highest, which is our WordFS-P-wP method, with a difference of 0.03 when reduced to 100. Both feature selection-based approaches perform better than the existing methods. As expected, the WordFS-S-wP method performs better than the WordFS-P-wP method in this task. Our WordFS-S-wP method outperforms the best performance from existing methods with an average improvement of 0.61, 1.97, and 3.69 in terms of Spearman's rank correlation coefficients when reducing to 150, 100, and 50 dimensions, respectively. Our method can perform better than the original 300-dimensional word embeddings, even when reduced to 100 dimensions.
For the dimension-reduced vectors based on the Word2vec embeddings, applying PPA may decrease performance, as the word embeddings are pre-trained on larger data, containing subtle linguistic connections. Our method can perform highest when the word vector dimension is reduced to 150 and 100 and slightly lower when reduced to 50. For the Fasttext word embeddings, although our method may not be the best, it consistently performs very close to the highest performance.

Then, we aggregated the twelve word similarity datasets. We first scaled them into the same range [0, 1] and then averaged the similarity scores of overlapped word pairs. There are two reasons to aggregate the datasets: 1) to provide an overall evaluation for the word similarity tasks and  2) to construct a comprehensive dataset to train our feature selection methods for various downstream tasks. We refer to our model as a weakly-supervised method for downstream tasks because there are only 7,705 human-labeled similarity scores for different word pairs in the aggregated dataset, which is much less than the total number of possible word pairs in the corpus to pre-train the word embeddings. For example, there are 400K word vectors in the pre-trained Glove word embeddings, which can generate approximately 80 billion unique word pairs. The other two word embeddings we use in the experiments have an even more extensive vocabulary. Compared with the possible word pairs, the guidance provided by the aggregated dataset is limited. We experimented with the aggregated dataset using the same settings as before.

\begin{table}[t]
\begin{center}
\caption{\label{agg_word_similarity}
Performance comparison of Spearman's rank correlation coefficients on the aggregated word similarity dataset.
}
\small
\setlength{\tabcolsep}{6pt} 
\renewcommand{\arraystretch}{1.2} 
\begin{tabular}{c|c|c|c|c}
\hline
\textbf{Dimension} & \textbf{300} & \textbf{150} & \textbf{100} & \textbf{50}\\
\hline
\text{Glove} & \text{45.74} & \text{--} & \text{38.45} & \text{36.24}\\
\text{UMAP} & \text{--} & \text{13.97} & \text{13.82} & \text{13.48}\\
\text{Algo} & \text{--} & \text{53.48} & \text{49.35} & \text{42.61}\\
\text{EmbedTextNet} & \text{--} & \text{52.71} & \text{43.88} & \text{41.00}\\
\text{WordFS-P-woP(ours)} & \text{--} & \text{48.23} & \text{48.39} & \text{48.22}\\
\text{WordFS-P-wP(ours)} & \text{--} & \text{54.69} & \text{52.60} & \text{47.29}\\
\text{WordFS-S-woP(ours)} & \text{--} & \underline{55.85} & \underline{54.45} & \underline{49.32}\\
\text{WordFS-S-wP(ours)} & \text{--} & \textbf{56.76} & \textbf{54.73} & \textbf{50.11}\\
\hline
\text{Word2vec} & \text{59.51} & \text{--} & \text{--} & \text{--}\\
\text{UMAP} & \text{--} & \text{32.54} & \text{32.20} & \text{32.16}\\
\text{Algo} & \text{--} & \text{59.11} & \text{56.98} & \text{50.74}\\
\text{EmbedTextNet} & \text{--} & \text{58.81} & \text{57.53} & \underline{54.15}\\
\text{WordFS-P-woP(ours)} & \text{--} & \text{59.06} & \text{57.37} & \text{53.58}\\
\text{WordFS-P-wP(ours)} & \text{--} & \text{59.15} & \text{57.20} & \text{52.99}\\
\text{WordFS-S-woP(ours)} & \text{--} & \underline{59.53} & \underline{58.04} & \textbf{54.82}\\
\text{WordFS-S-wP(ours)} & \text{--} & \textbf{59.76} & \textbf{58.26} & \text{54.13}\\
\hline
\text{Fasttext} & \text{54.63} & \text{--} & \text{--} & \text{--}\\
\text{UMAP} & \text{--} & \text{38.48} & \text{38.54} & \text{38.40}\\
\text{Algo} & \text{--} & \text{58.24} & \text{56.12} & \text{49.76}\\
\text{EmbedTextNet} & \text{--} & \text{56.78} & \text{55.11} & \text{51.17}\\
\text{WordFS-P-woP(ours)} & \text{--} & \text{52.16} & \text{46.55} & \text{35.49}\\
\text{WordFS-P-wP(ours)} & \text{--} & \text{60.86} & \text{57.69} & \text{50.46}\\
\text{WordFS-S-woP(ours)} & \text{--} & \underline{64.22} & \underline{61.65} & \underline{56.87}\\
\text{WordFS-S-wP(ours)} & \text{--} & \textbf{65.18} & \textbf{63.54} & \textbf{57.43}\\
\hline
\end{tabular}
\end{center}
\end{table}

Table~\ref{agg_word_similarity} shows the results of various methods for dimension reduction on the aggregated dataset. The annotations in the tables are the same as in the previous one. The performance of UMAP is still very low. Our WordFS-S methods perform best among all the experiments with different pre-trained word embeddings in other dimensions. And there is a large gap between our methods and the existing methods. Our approaches significantly benefit this task compared to the PCA-based method, particularly in lower dimensions where the gap widens. The performance of the EmbedTextNet model improves at lower dimensions but still lags behind our WordFS-S methods. It is worth noting that even when reduced to 50 dimensions, our WordFS-S-wP method achieves better performance compared to the original 300-dimensional word embeddings. Our experiment with three different pre-trained word embeddings shows that our method achieved the best performance for the Glove and Fasttext embeddings when post-processing is applied. However, for the Word2vec embeddings, better results are obtained without post-processing. This is reasonable because the power of the post-processing method may depend on the pre-trained word embeddings and the specific application scenario. The Word2vec embeddings we used were pre-trained on the most extensive data among the three. It implies that the embeddings may be more resilient to noise and capable of capturing more nuanced information. However, applying PPA can potentially result in the loss of information and cause a drop in performance.

\subsection{Downstream Tasks}
It is insufficient to only evaluate word embeddings based on word similarity tasks \cite{faruqui2016problems}. To demonstrate the power of our method, generalizability on downstream tasks is crucial. We utilize the SentEval toolkit \cite{conneau2018senteval} to evaluate the dimension-reduced vectors on various downstream tasks. We conducted experiments on nine prediction tasks and five sentence similarity tasks. The sentence representations were obtained by averaging the word embeddings. We utilize the WordFS-P for the prediction tasks and the WordFS-S method for the sentence similarity tasks.

\textbf{Prediction tasks:}
We conduct experiments on nine prediction tasks from the SentEval toolkit, including binary and multi-class classification tasks (MR, CR, MPQA, SUBJ, STS-B, STS-FG, and TREC), paraphrase detection task (MRPC), and entailment and semantic relatedness task (SICK-E). These tasks directly take the word embeddings as input. We applied our method with the RFT feature selection based on weak guidance from the aggregated word similarity dataset to the pre-trained word embeddings in all the experiments to reduce the dimension. Then, we directly input the dimension-reduced embeddings to the downstream tasks and compare the test accuracy of different methods for each task.

\begin{table}
\begin{center}
\caption{\label{prediction_tasks}
Comparison of the results of downstream prediction tasks, where the last column is the average of the previous columns.
}
\setlength{\tabcolsep}{7pt} 
\renewcommand{\arraystretch}{1.2} 
\resizebox*{\linewidth}{!}{
\begin{tabular}{c|c|c|c|c|c|c|c|c|c|c}
\noalign{\hrule height 2pt}
 \textbf{Task} & \textbf{MR} & \textbf{CR} & \textbf{MPQA} & \textbf{SUBJ} & \textbf{STS-B} & \textbf{SST} & \textbf{TREC} & \textbf{MRPC} & \textbf{SICK-E} & \textbf{Avg}\\
 &  &  &  &  &  & \textbf{-FG} &  &  &  &\\
\noalign{\hrule height 2pt}
\text{Glove-300D} & \text{74.99} & \text{75.81} & \text{86.75} & \text{91.39} & \text{78.20} & \text{40.77} & \text{66.6} & \text{72.58} & \text{77.19} & \text{73.81} \\
\hline
\text{UMAP-150D} & \text{50.00} & \text{63.76} & \text{68.77} & \text{50.00} & \text{49.92} & \text{28.64} & \text{18.8} & \text{67.48} & \text{56.69} & \text{50.45} \\
\text{Algo-150D} & \underline{73.70} & \underline{75.32} & \text{85.31} & \text{89.59} & \textbf{76.66} & \underline{40.95} & \text{60.7} & \textbf{71.68} & \underline{76.49} & \text{72.27} \\
\text{EmbedTextNet-150D} & \text{72.67} & \text{74.09} & \text{84.95} & \textbf{90.08} & \text{73.75} & \text{40.50} & \text{64.8} & \underline{71.54} & \text{74.43} & \text{71.87} \\
\text{WordFS-P-woP-150D(ours)} & \text{73.09} & \text{74.86} & \textbf{86.24} & \underline{89.86} & \text{76.22} & \text{39.55} & \textbf{65.4} & \text{71.01} & \textbf{76.62} & \underline{72.54} \\
\text{WordFS-P-wP-150D(ours)} & \textbf{74.03} & \textbf{76.00} & \textbf{86.24} & \text{89.20} & \underline{76.50} & \textbf{41.27} & \underline{65.2} & \text{70.38} & \text{76.46} & \textbf{72.81} \\
\hline
\text{UMAP-100D} & \text{50.00} & \text{63.82} & \text{68.77} & \text{50.00} & \text{49.92} & \text{28.64} & \text{18.8} & \text{67.36} & \text{56.69} & \text{50.44} \\
\text{Algo-100D} & \text{70.61} & \text{72.82} & \text{82.71} & \text{88.16} & \underline{74.63} & \text{38.28} & \text{56.2} & \textbf{71.83} & \underline{75.36} & \text{70.07} \\
\text{EmbedTextNet-100D} & \text{71.55} & \text{72.87} & \text{84.44} & \textbf{88.88} & \text{70.18} & \underline{38.46} & \underline{62.2} & \text{71.48} & \text{75.14} & \text{70.58} \\
\text{WordFS-P-woP-100D(ours)} & \underline{72.24} & \underline{73.96} & \underline{85.26} & \underline{88.62} & \text{73.97} & \text{38.37} & \text{62.0} & \text{71.42} & \textbf{76.60} & \underline{71.38} \\
\text{WordFS-P-wP-100D(ours)} & \textbf{72.68} & \textbf{75.42} & \textbf{85.50} & \text{88.25} & \textbf{76.94} & \textbf{40.50} & \textbf{64.4} & \underline{71.59} & \text{74.87} & \textbf{72.24} \\
\hline
\text{UMAP-50D} & \text{50.00} & \text{63.76} & \text{68.78} & \text{50.00} & \text{49.92} & \text{28.64} & \text{18.8} & \text{67.83} & \text{56.69} & \text{50.49} \\
\text{Algo-50D} & \text{66.24} & \text{70.23} & \text{76.35} & \text{84.32} & \text{69.36} & \text{35.88} & \text{48.1} & \underline{71.39} & \text{72.98} & \text{66.09} \\
\text{EmbedTextNet-50D} & \underline{69.36} & \textbf{72.34} & \underline{82.79} & \textbf{87.65} & \underline{71.66} & \textbf{38.64} & \textbf{60.2} & \text{68.29} & \textbf{74.20} & \textbf{69.45} \\
\text{WordFS-P-woP-50D(ours)} & \text{68.54} & \text{70.54} & \text{81.35} & \underline{85.51} & \text{69.85} & \text{35.61} & \text{56.2} & \textbf{71.65} & \underline{73.19} & \text{68.05} \\
\text{WordFS-P-wP-50D(ours)} & \textbf{69.96} & \underline{71.05} & \textbf{83.29} & \text{84.25} & \textbf{72.10} & \underline{37.69} & \underline{56.6} & \text{71.30} & \text{72.82} & \underline{68.78} \\
\noalign{\hrule height 2pt}
\text{Word2vec-300D} & \text{77.16} & \text{77.80} & \text{87.97} & \text{90.42} & \text{81.11} & \text{42.22} & \text{82.6} & \text{72.35} & \text{77.92} & \text{76.62} \\
\hline
\text{UMAP-150D} & \text{50.03} & \text{63.76} & \text{68.74} & \text{50.00} & \text{49.92} & \text{28.64} & \text{18.8} & \text{66.49} & \text{56.69} & \text{50.34} \\
\text{Algo-150D} & \underline{74.52} & \text{75.28} & \text{85.72} & \underline{89.31} & \underline{77.70} & \text{39.46} & \text{70.0} & \text{70.55} & \text{73.13} & \text{72.85} \\
\text{EmbedTextNet-150D} & \text{73.87} & \underline{77.14} & \text{84.98} & \textbf{89.73} & \text{77.05} & \textbf{40.72} & \underline{76.0} & \textbf{73.28} & \text{73.88} & \text{74.07} \\
\text{WordFS-P-woP-150D(ours)} & \text{74.43} & \text{76.90} & \textbf{86.56} & \text{88.72} & \textbf{79.13} & \underline{40.18} & \textbf{77.0} & \text{72.12} & \textbf{76.54} & \textbf{74.65} \\
\text{WordFS-P-wP-150D(ours)} & \textbf{75.51} & \textbf{77.41} & \underline{86.15} & \text{88.45} & \text{76.88} & \text{40.14} & \text{73.8} & \underline{72.17} & \textbf{76.54} & \underline{74.12} \\
\hline
\text{UMAP-100D} & \text{50.00} & \text{63.76} & \text{68.77} & \text{50.00} & \text{53.71} & \text{28.64} & \text{18.8} & \text{66.49} & \text{58.78} & \text{50.99} \\
\text{Algo-100D} & \text{72.10} & \text{74.70} & \text{84.42} & \underline{88.30} & \underline{76.11} & \text{38.60} & \text{62.2} & \textbf{72.87} & \underline{74.08} & \text{71.49} \\
\text{EmbedTextNet-100D} & \textbf{72.93} & \textbf{76.40} & \textbf{85.43} & \textbf{89.33} & \textbf{77.81} & \text{36.65} & \text{65.8} & \text{72.29} & \text{73.78} & \text{72.27} \\
\text{WordFS-P-woP-100D(ours)} & \underline{72.75} & \text{74.25} & \underline{84.77} & \text{87.77} & \text{76.06} & \underline{39.77} & \textbf{75.2} & \underline{72.70} & \text{73.68} & \textbf{72.97} \\
\text{WordFS-P-wP-100D(ours)} & \underline{72.75} & \underline{74.83} & \text{84.65} & \text{86.52} & \text{75.78} & \textbf{40.05} & \underline{73.0} & \text{70.09} & \textbf{75.24} & \underline{72.54} \\
\hline
\text{UMAP-50D} & \text{50.00} & \text{63.76} & \text{68.70} & \text{50.00} & \text{52.50} & \text{28.64} & \text{13.00} & \text{66.49} & \text{56.69} & \text{49.98} \\
\text{Algo-50D} & \underline{70.39} & \text{70.84} & \underline{83.39} & \underline{86.73} & \underline{73.64} & \text{37.06} & \text{54.4} & \textbf{72.70} & \text{69.60} & \text{68.75} \\
\text{EmbedTextNet-50D} & \textbf{70.79} & \textbf{72.95} & \textbf{84.35} & \textbf{87.47} & \textbf{74.74} & \textbf{40.00} & \textbf{66.8} & \text{70.84} & \textbf{72.21} & \textbf{71.13} \\
\text{WordFS-P-woP-50D(ours)} & \text{69.39} & \underline{71.60} & \text{81.35} & \text{85.54} & \text{72.71} & \text{37.69} & \underline{62.6} & \underline{72.52} & \text{70.90} & \underline{69.37} \\
\text{WordFS-P-wP-50D(ours)} & \text{70.02} & \text{71.52} & \text{82.48} & \text{82.66} & \text{72.43} & \underline{38.10} & \text{61.6} & \text{71.65} & \underline{71.10} & \text{69.06} \\
\noalign{\hrule height 2pt}
\text{Fasttext-300D} & \text{77.59} & \text{78.81} & \text{88.10} & \text{91.73} & \text{80.78} & \text{45.48} & \text{84.4} & \text{72.99} & \text{77.57} & \text{77.49} \\
\hline
\text{UMAP-150D} & \text{50.00} & \text{63.76} & \text{68.80} & \text{50.00} & \text{49.92} & \text{28.64} & \text{18.8} & \text{71.07} & \text{56.69} & \text{50.85} \\
\text{Algo-150D} & \text{74.81} & \text{75.39} & \text{85.72} & \textbf{90.12} & \text{77.81} & \text{41.58} & \text{66.4} & \textbf{73.39} & \text{70.79} & \text{72.89} \\
\text{EmbedTextNet-150D} & \text{73.95} & \textbf{76.77} & \textbf{86.32} & \underline{90.80} & \text{78.91} & \text{40.54} & \underline{73.2} & \text{72.12} & \underline{74.79} & \text{74.16} \\
\text{WordFS-P-woP-150D(ours)} & \underline{75.68} & \underline{75.89} & \textbf{86.32} & \text{89.62} & \textbf{79.68} & \underline{42.04} & \textbf{76.0} & \underline{72.29} & \text{74.24} & \textbf{74.64} \\
\text{WordFS-P-wP-150D(ours)} & \textbf{76.01} & \text{75.87} & \text{86.13} & \text{89.39} & \underline{79.57} & \textbf{42.08} & \text{73.0} & \text{72.12} & \textbf{75.28} & \underline{74.38} \\
\hline
\text{UMAP-100D} & \text{50.00} & \text{63.76} & \text{68.77} & \text{50.00} & \text{49.92} & \text{28.64} & \text{16.2} & \text{66.49} & \text{56.69} & \text{50.05} \\
\text{Algo-100D} & \text{73.31} & \underline{76.05} & \underline{85.43} & \underline{88.96} & \underline{77.27} & \text{40.50} & \text{66.0} & \underline{72.58} & \text{70.98} & \text{72.34} \\
\text{EmbedTextNet-100D} & \textbf{74.56} & \textbf{76.74} & \textbf{86.17} & \textbf{90.09} & \textbf{77.87} & \textbf{41.22} & \textbf{74.8} & \text{72.52} & \textbf{74.41} & \textbf{74.26} \\
\text{WordFS-P-woP-100D(ours)} & \underline{74.43} & \text{74.86} & \text{84.16} & \text{88.54} & \text{76.94} & \text{40.23} & \underline{69.8} & \text{72.06} & \underline{71.79} & \underline{72.53} \\
\text{WordFS-P-wP-100D(ours)} & \text{73.51} & \text{73.56} & \text{85.06} & \text{86.33} & \text{76.83} & \underline{40.54} & \text{69.4} & \textbf{73.04} & \text{71.34} & \text{72.18} \\
\hline
\text{UMAP-50D} & \text{50.04} & \text{63.76} & \text{68.79} & \text{50.00} & \text{49.92} & \text{28.64} & \text{32.6} & \text{66.90} & \text{56.69} & \text{51.93} \\
\text{Algo-50D} & \text{70.99} & \underline{71.55} & \underline{84.12} & \underline{86.14} & \underline{73.09} & \textbf{38.14} & \text{55.0} & \underline{72.64} & \text{70.47} & \underline{69.13} \\
\text{EmbedTextNet-50D} & \textbf{72.34} & \textbf{72.08} & \textbf{84.85} & \textbf{87.97} & \text{72.60} & \text{37.65} & \textbf{72.6} & \textbf{73.16} & \textbf{71.40} & \textbf{71.63} \\
\text{WordFS-P-woP-50D(ours)} & \underline{71.82} & \text{70.33} & \text{81.43} & \text{83.98} & \textbf{74.63} & \underline{37.87} & \underline{60.0} & \text{71.71} & \text{70.20} & \text{69.11} \\
\text{WordFS-P-wP-50D(ours)} & \text{70.23} & \text{69.11} & \text{81.41} & \text{81.79} & \text{71.77} & \text{36.61} & \text{57.0} & \text{71.94} & \underline{70.51} & \text{67.82} \\
\noalign{\hrule height 2pt}
\end{tabular}
}
\end{center}
\end{table}

\begin{figure}[h]
\centering
\includegraphics[width=1\textwidth]{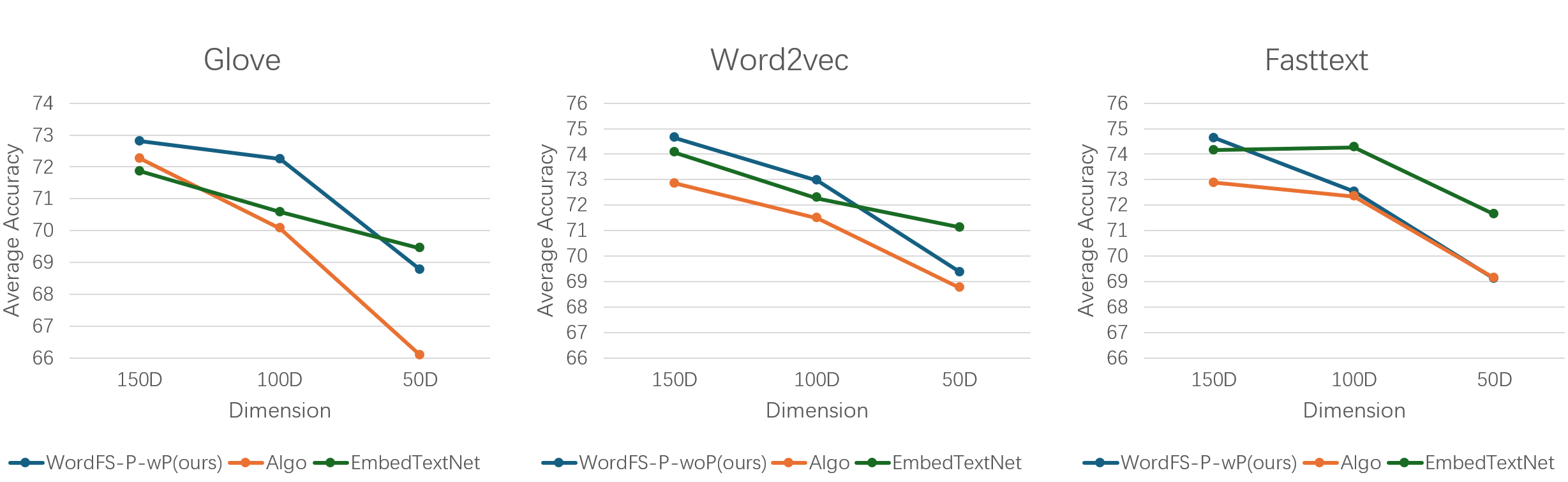}
\caption{Average accuracy comparison for prediction tasks.}\label{fig:prediction_tasks}
\end{figure}

Table~\ref{prediction_tasks} shows the results of various methods on the nine prediction tasks. The last column is the average of the previous columns. Figure \ref{fig:prediction_tasks} compares the average accuracy among the top three methods. Experiments show that our method surpasses the UMAP method in all the prediction tasks and outperforms the PCA-based method in most tasks. Also, our method achieves higher average scores than the PCA-based method in all the settings, except for a slightly lower performance when reducing the Fasttext word embedding from 300 to 50 dimensions, with a difference of only 0.02. Our method also performs better than the EmbedTextNet model, although it performs slightly worse when reduced to 50 dimensions. However, as shown in the next section, our model is much more efficient and takes less time to train than the EmbedTextNet model. Our method retains more helpful information for most settings while reducing dimensions, achieving a much closer prediction performance to the original 300-dimensional word embeddings. It indicates the effectiveness of our method. Also, it proves that our method can generalize well on various prediction downstream tasks in a zero-shot manner. The Glove word embedding we used was pre-trained on a smaller amount of data and might be vulnerable to noise. Post-processing can help mitigate noise and improve the performance. However, subtle linguistic connections can be crucial when working with prediction tasks. Applying PPA to word embeddings pre-trained on relatively large data may decrease performance.

\begin{figure}[h]
\centering
\includegraphics[width=1\textwidth]{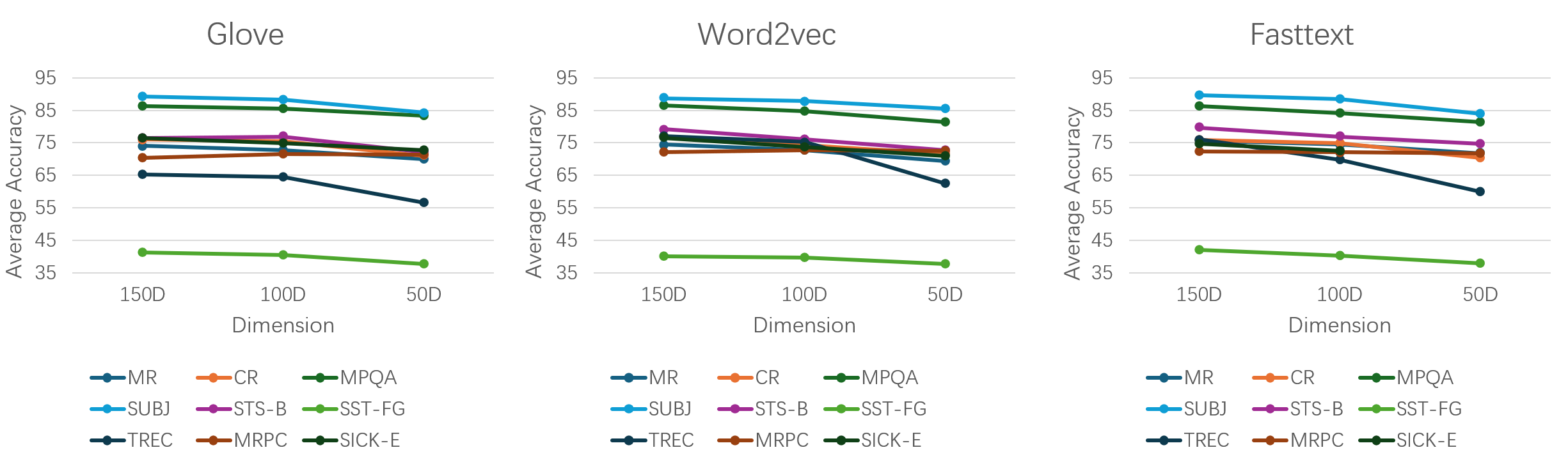}
\caption{Accuracy comparison for prediction tasks using our WordFS method.}\label{fig:all_tasks_pred}
\end{figure}

Reducing the dimension of data results in information loss, often leading to decreased performance as the dimension decreases. Figure \ref{fig:all_tasks_pred} compares the impact of word vector dimensions on performance across different tasks. The impact of dimension reduction on task performance can vary depending on the pre-trained word embeddings used, but the overall trend remains consistent. Various tasks may lose crucial information at different reduction levels. For example, the TREC task's performance significantly drops when reduced to 50D compared to 100D, while the performance of most other tasks decreases gradually when reduced to a lower amount. When using Word2vec and Fasttext word embeddings, the TREC task's performance drops significantly when the dimension is reduced to 150D. This suggests that important information about the type of question is lost early on. For most other tasks, performance decreases slightly as the dimension decreases, which is expected when dimensions are reduced. In conclusion, dimension reduction models can preserve essential information for improved overall performance, but the performance of specific tasks may vary without further fine-tuning.

\textbf{Similarity tasks:}
Furthermore, we evaluate our method on five sentence similarity tasks from the SentEval toolkit. The sentence similarity tasks contain STS tasks from 2012 to 2016, aiming to determine how close the distance between two sentence vectors correlates with a similarity score assigned by human annotators. We first applied our Sim-based method to reduce the word embeddings based on the weak guidance provided by the aggregated word similarity dataset. Then, we directly input the dimension-reduced vectors and compare the average Spearman's rank correlation coefficients among different methods.

Table~\ref{STS_tasks} shows the results of various methods on the sentence similarity tasks. Figure \ref{fig:prediction_tasks} compares the average Spearman's rank correlation coefficients among different methods. Our method is consistently better than the existing methods, except when we reduce Glove embedding to 150 dimensions and Word2vec embedding to 50 dimensions, where the existing methods perform better with less than 0.05 difference. Our method significantly outperforms the existing approach in most sentence similarity tasks, with a gap of up to 8.36. The observation of post-processing is the same as that of the aggregated word similarity dataset. The Word2vec embeddings we use are trained on extensive data and may not benefit much from the PPA. The performance of the reduced word embeddings is often as good or even better than the original word vectors for Glove and Fasttext word embeddings in sentence similarity tasks. Even for Word2vec word embeddings, the drop in performance is small. This indicates that our method, which is based on word similarity, effectively retains critical dimensions for sentence similarity tasks while reducing the confusing ones.

\begin{table}
\begin{center}
\caption{\label{STS_tasks}
Performance comparison of downstream sentence similarity tasks, where the last column is the average of the previous columns.
}
\scriptsize
\setlength{\tabcolsep}{1pt} 
\renewcommand{\arraystretch}{1.4} 
\begin{tabular}{c|c|c|c|c|c|c}
\noalign{\hrule height 2pt}
\textbf{Task} & \textbf{STS} & \textbf{STS} & \textbf{STS} & \textbf{STS} &\textbf{STS} & \textbf{Avg}\\
& \textbf{-12} & \textbf{-13} & \textbf{-14} & \textbf{-15} & \textbf{-16} \\
\noalign{\hrule height 2pt}
\text{Glove-300D} & \text{51.52} & \text{48.40} & \text{54.21} & \text{57.09} & \text{55.28} & \text{53.30}\\
\hline
\text{UMAP-150D} & \text{39.10} & \text{26.16} & \text{28.96} & \text{33.65} & \text{40.94} & \text{33.76}\\
\text{Algo-150D} & \textbf{53.23} & \underline{56.41} & \textbf{62.30} & \textbf{67.52} & \textbf{66.99} & \textbf{61.29}\\
\text{EmbedTextNet-150D} & \text{52.68} & \text{52.12} & \text{58.67} & \text{61.87} & \text{59.80} & \text{57.03}\\
\text{WordFS-S-woP-150D(ours)} & \text{51.40} & \text{53.75} & \text{59.19} & \text{63.07} & \text{62.16} & \text{57.91}\\
\text{WordFS-S-wP-150D(ours)} & \underline{52.74} & \textbf{58.62} & \underline{62.16} & \underline{66.42} & \underline{66.24} & \underline{61.24}\\
\hline
\text{UMAP-100D} & \text{39.69} & \text{26.90} & \text{28.97} & \text{33.98} & \text{41.40} & \text{26.79}\\
\text{Algo-100D} & \textbf{52.51} & \underline{54.66} & \underline{60.82} & \underline{66.79} & \underline{65.15} & \underline{59.99}\\
\text{EmbedTextNet-100D} & \text{51.89} & \text{48.30} & \text{55.51} & \text{58.43} & \text{55.66} & \text{53.96}\\
\text{WordFS-S-woP-100D(ours)} & \text{49.83} & \text{51.00} & \text{57.98} & \text{61.59} & \text{62.74} & \text{56.63}\\
\text{WordFS-S-wP-100D(ours)} & \underline{52.16} & \textbf{59.05} & \textbf{62.24} & \textbf{66.84} & \textbf{66.21} & \textbf{61.30}\\
\hline
\text{UMAP-50D} & \text{38.19} & \text{26.92} & \text{29.17} & \text{33.52} & \text{39.94} & \text{33.55}\\
\text{{Algo-50D}} & \text{48.20} & \text{49.44} & \underline{57.53} & \underline{61.49} & \underline{61.42} & \underline{55.62}\\
\text{EmbedTextNet-50D} & \text{48.83} & \text{45.86} & \text{53.57} & \text{56.80} & \text{53.32} & \text{51.68}\\
\text{WordFS-S-woP-50D(ours)} & \textbf{50.27} & \underline{49.58} & \text{55.66} & \text{56.96} & \text{60.07} & \text{54.51}\\
\text{WordFS-S-wP-50D(ours)} & \underline{49.71} & \textbf{55.42} & \textbf{60.58} & \textbf{63.50} & \textbf{63.62} & \textbf{58.57}\\
\noalign{\hrule height 2pt}
\text{Word2vec-300D} & \text{55.48} & \text{58.23} & \text{64.05} & \text{67.97} & \text{66.29} & \text{62.40}\\
\hline
\text{UMAP-150D} & \text{38.01} & \text{33.32} & \text{37.75} & \text{41.73} & \text{46.29} & \text{39.72}\\
\text{Algo-150D} & \underline{54.94} & \text{58.63} & \text{64.01} & \text{65.90} & \text{65.14} & \text{61.72}\\
\text{EmbedTextNet-150D} & \text{54.85} & \underline{58.91} & \textbf{64.31} & \text{66.79} & \text{65.45} & \text{62.06}\\
\text{WordFS-S-woP-150D(ours)} & \text{54.91} & \textbf{59.02} & \text{64.07} & \textbf{67.70} & \underline{66.68} & \textbf{62.48}\\
\text{WordFS-S-wP-150D(ours)} & \textbf{55.12} & \text{58.90} & \underline{64.30} & \underline{67.13} & \textbf{66.73} & \underline{62.44}\\
\hline
\text{UMAP-100D} & \text{37.87} & \text{35.87} & \text{38.15} & \text{42.11} & \text{46.92} & \text{40.18}\\
\text{Algo-100D} & \text{54.10} & \text{57.43} & \underline{63.54} & \text{64.58} & \text{63.31} & \text{60.59}\\
\text{EmbedTextNet-100D} & \text{54.94} & \text{57.68} & \textbf{63.79} & \text{65.01} & \text{64.96} & \text{61.28}\\
\text{WordFS-S-woP-100D(ours)} & \textbf{55.38} & \underline{57.77} & \text{63.32} & \textbf{67.20} & \textbf{66.27} & \textbf{61.99}\\
\text{WordFS-S-wP-100D(ours)} & \underline{54.96} & \textbf{58.33} & \text{63.36} & \underline{67.04} & \underline{65.97} & \underline{61.93}\\
\hline
\text{UMAP-50D} & \text{38.12} & \text{31.41} & \text{37.00} & \text{41.37} & \text{46.75} & \text{38.93}\\
\text{{Algo-50D}} & \text{51.20} & \text{53.46} & \text{60.51} & \text{61.89} & \text{60.76} & \text{57.56}\\
\text{EmbedTextNet-50D} & \textbf{54.23} & \textbf{55.85} & \underline{61.81} & \text{62.64} & \text{62.69} & \textbf{59.44}\\
\text{WordFS-S-woP-50D(ours)} & \text{52.70} & \text{55.45} & \text{61.55} & \textbf{64.14} & \textbf{62.93} & \text{59.35}\\
\text{WordFS-S-wP-50D(ours)} & \underline{53.03} & \underline{55.61} & \textbf{61.99} & \underline{63.64} & \underline{62.83} & \underline{59.42}\\
\noalign{\hrule height 2pt}
\text{Fasttext-300D} & \text{56.15} & \text{51.34} & \text{59.11} & \text{63.21} & \text{61.03} & \text{58.17}\\
\hline
\text{UMAP-150D} & \text{42.52} & \text{29.34} & \text{36.48} & \text{37.06} & \text{44.37} & \text{37.95}\\
\text{Algo-150D} & \text{56.80} & \text{55.03} & \text{61.72} & \text{64.07} & \text{61.05} & \text{59.73}\\
\text{EmbedTextNet-150D} & \text{55.22} & \text{51.03} & \text{59.55} & \text{62.67} & \text{60.94} & \text{57.88}\\
\text{WordFS-S-woP-150D(ours)} & \underline{59.05} & \underline{56.76} & \underline{62.73} & \underline{67.33} & \underline{64.63} & \underline{62.10}\\
\text{WordFS-S-wP-150D(ours)} & \textbf{60.72} & \textbf{62.79} & \textbf{64.82} & \textbf{69.95} & \textbf{70.02} & \textbf{65.66}\\
\hline
\text{UMAP-100D} & \text{41.49} & \text{28.76} & \text{36.17} & \text{37.09} & \text{43.99} & \text{37.50}\\
\text{Algo-100D} & \text{56.57} & \underline{54.43} & \text{60.06} & \text{62.53} & \text{60.09} & \text{58.74}\\
\text{EmbedTextNet-100D} & \text{55.55} & \text{48.91} & \text{58.48} & \text{61.42} & \text{58.86} & \text{56.64}\\
\text{WordFS-S-woP-100D(ours)} & \underline{57.74} & \text{53.78} & \underline{61.07} & \underline{65.60} & \underline{62.37} & \underline{60.11} \\
\text{WordFS-S-wP-100D(ours)} & \textbf{60.25} & \textbf{60.84} & \textbf{64.21} & \textbf{69.54} & \textbf{69.14} & \textbf{64.80}\\
\hline
\text{UMAP-50D} & \text{42.25} & \text{28.16} & \text{36.01} & \text{37.20} & \text{44.43} & \text{37.61}\\
\text{Algo-50D} & \underline{55.56} & \underline{53.82} & \text{59.99} & \text{60.60} & \textbf{68.91} & \text{57.78}\\
\text{EmbedTextNet-50D} & \text{55.34} & \text{49.76} & \text{58.98} & \text{59.70} & \text{58.52} & \text{56.46}\\
\text{WordFS-S-woP-50D(ours)} & \text{54.88} & \text{50.91} & \underline{60.44} & \underline{63.60} & \text{60.74} & \underline{58.11}\\
\text{WordFS-S-wP-50D(ours)} & \textbf{57.33} & \textbf{56.74} & \textbf{60.94} & \textbf{65.41} & \underline{64.10} & \textbf{60.90}\\
\noalign{\hrule height 2pt}
\end{tabular}
\end{center}
\end{table}

\begin{figure}[t]
\centering
\includegraphics[width=1\textwidth]{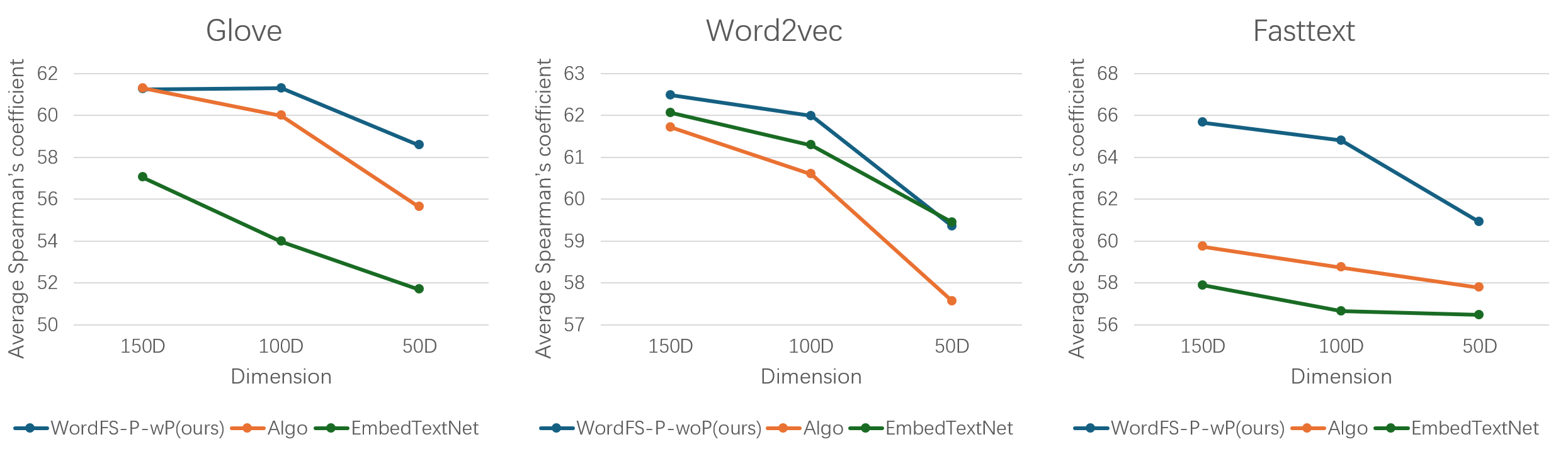}
\caption{Comparison of the average Spearman's rank correlation coefficients of sentence similarity tasks.}\label{fig:sent_tasks}
\end{figure}

\begin{figure}[h]
\centering
\includegraphics[width=1\textwidth]{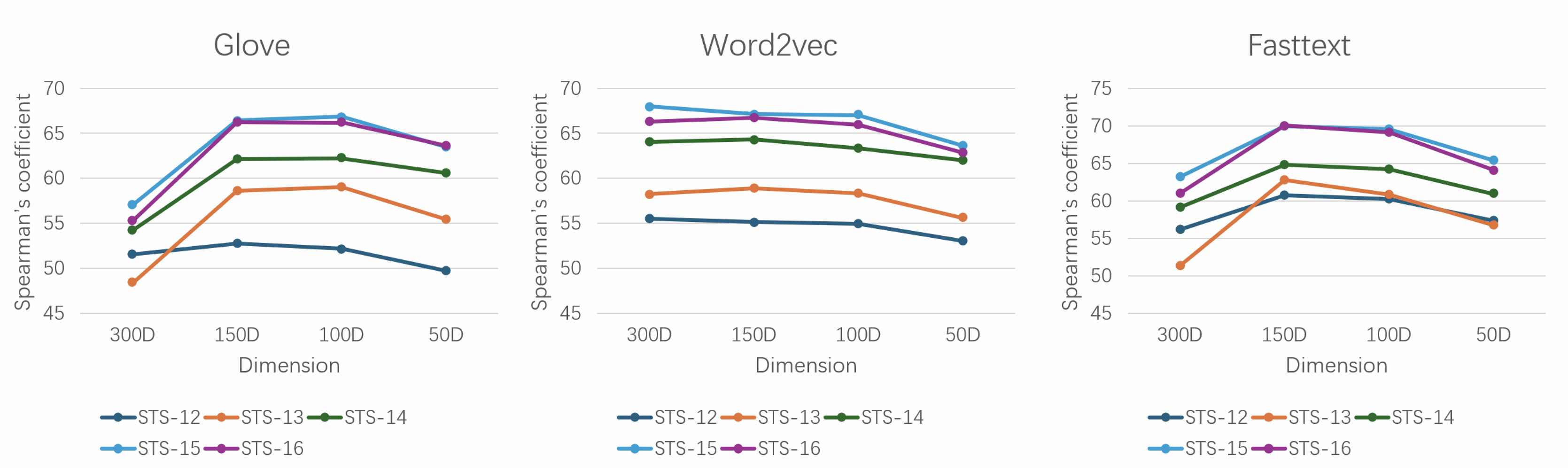}
\caption{Spearman's rank correlation coefficients comparison for similarity tasks using our WordFS method.}\label{fig:all_tasks_sim}
\end{figure}

Figure \ref{fig:all_tasks_sim} compares the impact of word vector dimensions on performance across different tasks. As the dimensions decrease, there is generally a slight drop in performance because some detailed information is discarded. Most lines show a similar trend from 150D to 50D. For the Glove and Word2vec embeddings, our approach significantly improves performance when reducing to 150D. This is understandable because these embeddings are trained on a smaller amount of data, and our approach helps the vectors focus on similarity tasks. Word2vec word embeddings are trained on a much larger amount of data, which is sufficient for them to perform well on sentence similarity tasks.

In conclusion, our method generally outperforms existing methods in prediction and similarity tasks, demonstrating its superiority and generalizability.

\subsection{Model Efficiency}
We compare the model efficiency of the Algo method, the EmbedTextNet method, and our method when reducing the word embedding dimensions from 300 to 150 in Table \ref{comp_complexity}. For each pre-trained word embedding, all methods start with the same vocabulary of 300D word vectors and output a file with 150D word vectors. Different hardware is used for other methods because of their algorithmic nature. All the experiments were conducted on the same server, equipped with two AMD EPYC 7543 32-core CPUs and seven NVIDIA RTX A6000 GPUs. The deep learning-based model is trained using the GPUs, while the Algo model and our methods are trained using only the CPUs. The EmbedTextNet model takes approximately 30 minutes to train on relatively small GloVe word embeddings.

In contrast, the training time for our WordFS-P-wP and WordFS-S-wP models is less than 19 seconds and 13 seconds, respectively. Our methods without post-processing require even less time — the training of the WordFS-P-woP model is completed within one second. For more significant pre-trained word embeddings, it takes hours to train the EmbedTextNet model. However, our WordFS-P-wP and WordFS-S-wP methods only require 1-2 minutes, making them hundreds of times faster. Our WordFS-P-woP model, without post-processing, consistently completes in around one second, making it thousands or ten thousand times faster than the deep learning-based method. Besides, our method is significantly faster and outperforms UMAP. Because finding the K nearest neighbors may take a long time in a high-dimensional word vector space.

Additionally, our methods are more efficient than the PCA-based method because they employ PPA before and after the PCA, while we only apply it once before feature selection. The PPA consumes a substantial portion of the training time compared to the feature selection module. Our model, without the post-processing, can achieve even less time. The reason is the post-processing and PCA must be done on the entire vocabulary. However, our feature selection method only focuses on a subset of words that appear in the aggregated word similarity dataset, making the procedure very fast. And the module's processing time will not be significantly affected by the vocabulary size. 

In conclusion, Our methods take only a fraction of the time compared to deep learning methods and less than half the time compared to existing PCA-based methods. The experimental results demonstrate that our methods are much more efficient than existing methods.

\begin{table}[h]
\begin{center}
\caption{\label{comp_complexity}
Complexity comparison of the Algo method, the EmbedTextNet method, and our methods when reducing the word embedding dimensions from 300 to 150 in hardware configuration and training time.
}
\small
\setlength{\tabcolsep}{6pt} 
\renewcommand{\arraystretch}{1.2} 
\begin{tabular}{c|c|c|c}
\noalign{\hrule height 2pt}
\textbf{Embeddings} & \textbf{Method} & \textbf{Hardware Configuration} & \textbf{Training Time (s)}\\
\noalign{\hrule height 2pt}
\multirow{6}{*}{\text{Glove}} & \text{EmbedTextNet} & \text{GPU} & \text{2149.04 (5372.60X)} \\
\cline{2-4}
& \text{UMAP} & \text{}  & \text{655.52 (1638.80X)} \\
& \text{Algo} & \text{}  & \text{21.50 (53.75X)} \\
& \text{WordFS-S-wP (ours)} & \text{} & \text{17.63 (44.08X)} \\
& \text{WordFS-P-wP (ours)} & \text{CPU} & \text{11.84 (29.60X)} \\
& \text{WordFS-S-woP (ours)} & \text{} & \text{4.05 (10.13X)} \\
& \text{WordFS-P-woP (ours)} & \text{} & \text{0.40 (1X)} \\
\noalign{\hrule height 2pt}
\multirow{6}{*}{\text{Word2vec}} & \text{EmbedTextNet} & \text{GPU} & \text{18272.62 (30970.54X)} \\
\cline{2-4}
& \text{UMAP} & \text{}  & \text{8470.38 (14356.58X)} \\
& \text{Algo} & \text{}  & \text{198.93 (337.17X)} \\
& \text{WordFS-S-wP (ours)} & \text{} & \text{93.88 (159.12X)} \\
& \text{WordFS-P-wP (ours)} & \text{CPU} & \text{85.93 (145.64X)} \\
& \text{WordFS-S-woP (ours)} & \text{} & \text{7.36 (12.47X)} \\
& \text{WordFS-P-woP (ours)} & \text{} & \text{0.59 (1X)} \\
\noalign{\hrule height 2pt}
\multirow{6}{*}{\text{Fasttext}} & \text{EmbedTextNet} & \text{GPU} & \text{5306.65 (3537.77X)} \\
\cline{2-4}
& \text{UMAP} & \text{}  & \text{2971.88 (1982.59X)} \\
& \text{Algo} & \text{}  & \text{74.98 (49.99X)} \\
& \text{WordFS-S-wP (ours)} & \text{} & \text{52.20 (34.80X)} \\
& \text{WordFS-P-wP (ours)} & \text{CPU} & \text{36.54 (24.36X)} \\
& \text{WordFS-S-woP (ours)} & \text{} & \text{15.24 (10.16X)} \\
& \text{WordFS-P-woP (ours)} & \text{} & \text{1.50 (1X)} \\
\noalign{\hrule height 2pt}
\end{tabular}
\end{center}
\end{table}

\section{Conclusion and Future Work} \label{sec:conclusion}
This paper proposes a general dimension reduction method called WordFS for pre-trained word embeddings with the post-processing algorithm (PPA) and simple yet effective feature selection methods based on weak supervision provided by a limited number of word similarity pairs. Our method is more straightforward, efficient, intuitive, and effective in most cases than the existing methods. Empirical results show that our method outperforms existing methods in word similarity tasks and generalizes well to various downstream tasks. Our model demonstrates clear advantages for tasks significantly related to word similarity, like STS tasks. Even for tasks that might not directly correlate with word similarity, such as classification tasks, our model performs better on average, showing our approach's generalizability. We demonstrate that our proposed weakly-supervised feature selection method can effectively reduce word embedding dimensions and generalize to many downstream tasks with much lower computational costs. In the future, we would like to explore the compression of word embeddings further. Since feature selection methods may result in information loss, there is still room for improvement in the performance of prediction tasks between the original word embeddings and the reduced word vectors. Additionally, suitable feature selection methods can be developed to compress pre-trained word vectors in specific domains.

\bibliographystyle{plain}
\renewcommand\refname{Reference}
\bibliography{main}

\begin{thebibliography}{10}

\bibitem{ait2023anisotropy}
Mira Ait-Saada and Mohamed Nadif.
\newblock Is anisotropy truly harmful? a case study on text clustering.
\newblock In {\em Proceedings of the 61st Annual Meeting of the Association for Computational Linguistics (Volume 2: Short Papers)}, pages 1194--1203, 2023.

\bibitem{bojanowski2016enriching}
Piotr Bojanowski, Edouard Grave, Armand Joulin, and Tomas Mikolov.
\newblock Enriching word vectors with subword information.
\newblock {\em arXiv preprint arXiv:1607.04606}, 2016.

\bibitem{conneau2018senteval}
Alexis Conneau and Douwe Kiela.
\newblock Senteval: An evaluation toolkit for universal sentence representations.
\newblock {\em arXiv preprint arXiv:1803.05449}, 2018.

\bibitem{devlin2018bert}
Jacob Devlin, Ming-Wei Chang, Kenton Lee, and Kristina Toutanova.
\newblock Bert: Pre-training of deep bidirectional transformers for language understanding.
\newblock {\em arXiv preprint arXiv:1810.04805}, 2018.

\bibitem{erenel2020new}
Zafer Erenel, Oluwatayomi~Rereloluwa Adegboye, and Huseyin Kusetogullari.
\newblock A new feature selection scheme for emotion recognition from text.
\newblock {\em Applied Sciences}, 10(15):5351, 2020.

\bibitem{faruqui2014improving}
Manaal Faruqui and Chris Dyer.
\newblock Improving vector space word representations using multilingual correlation.
\newblock In {\em Proceedings of the 14th Conference of the European Chapter of the Association for Computational Linguistics}, pages 462--471, 2014.

\bibitem{faruqui2016problems}
Manaal Faruqui, Yulia Tsvetkov, Pushpendre Rastogi, and Chris Dyer.
\newblock Problems with evaluation of word embeddings using word similarity tasks.
\newblock {\em arXiv preprint arXiv:1605.02276}, 2016.

\bibitem{hwang2023embedtextnet}
Dae~Yon Hwang, Bilal Taha, and Yaroslav Nechaev.
\newblock Embedtextnet: Dimension reduction with weighted reconstruction and correlation losses for efficient text embedding.
\newblock In {\em Findings of the Association for Computational Linguistics: ACL 2023}, pages 9863--9879, 2023.

\bibitem{jha2021geodesic}
Rishi Jha and Kai Mihata.
\newblock On geodesic distances and contextual embedding compression for text classification.
\newblock {\em arXiv preprint arXiv:2104.11295}, 2021.

\bibitem{jiao2019tinybert}
Xiaoqi Jiao, Yichun Yin, Lifeng Shang, Xin Jiang, Xiao Chen, Linlin Li, Fang Wang, and Qun Liu.
\newblock Tinybert: Distilling bert for natural language understanding.
\newblock {\em arXiv preprint arXiv:1909.10351}, 2019.

\bibitem{kim2020adaptive}
Yeachan Kim, Kang-Min Kim, and SangKeun Lee.
\newblock Adaptive compression of word embeddings.
\newblock In {\em Proceedings of the 58th annual meeting of the association for computational linguistics}, pages 3950--3959, 2020.

\bibitem{kuo2023green}
C-C~Jay Kuo and Azad~M Madni.
\newblock Green learning: Introduction, examples and outlook.
\newblock {\em Journal of Visual Communication and Image Representation}, 90:103685, 2023.

\bibitem{mcinnes2018umap}
Leland McInnes, John Healy, and James Melville.
\newblock Umap: Uniform manifold approximation and projection for dimension reduction.
\newblock {\em arXiv preprint arXiv:1802.03426}, 2018.

\bibitem{mikolov2013efficient}
Tomas Mikolov, Kai Chen, Greg Corrado, and Jeffrey Dean.
\newblock Efficient estimation of word representations in vector space.
\newblock {\em arXiv preprint arXiv:1301.3781}, 2013.

\bibitem{mikolov2018advances}
Tomas Mikolov, Edouard Grave, Piotr Bojanowski, Christian Puhrsch, and Armand Joulin.
\newblock Advances in pre-training distributed word representations.
\newblock In {\em Proceedings of the International Conference on Language Resources and Evaluation (LREC 2018)}, 2018.

\bibitem{mu2018allbutthetop}
Jiaqi Mu and Pramod Viswanath.
\newblock All-but-the-top: Simple and effective postprocessing for word representations.
\newblock In {\em International Conference on Learning Representations}, 2018.

\bibitem{naveed2023comprehensive}
Humza Naveed, Asad~Ullah Khan, Shi Qiu, Muhammad Saqib, Saeed Anwar, Muhammad Usman, Nick Barnes, and Ajmal Mian.
\newblock A comprehensive overview of large language models.
\newblock {\em arXiv preprint arXiv:2307.06435}, 2023.

\bibitem{pennington2014glove}
Jeffrey Pennington, Richard Socher, and Christopher~D Manning.
\newblock Glove: Global vectors for word representation.
\newblock In {\em Proceedings of the 2014 conference on empirical methods in natural language processing (EMNLP)}, pages 1532--1543, 2014.

\bibitem{peters2018deep}
Matthew~E. Peters, Mark Neumann, Mohit Iyyer, Matt Gardner, Christopher Clark, Kenton Lee, and Luke Zettlemoyer.
\newblock Deep contextualized word representations, 2018.

\bibitem{raunak2019effective}
Vikas Raunak, Vivek Gupta, and Florian Metze.
\newblock Effective dimensionality reduction for word embeddings.
\newblock In {\em Proceedings of the 4th Workshop on Representation Learning for NLP (RepL4NLP-2019)}, pages 235--243, 2019.

\bibitem{rui2016unsupervised}
Weikang Rui, Jinwen Liu, and Yawei Jia.
\newblock Unsupervised feature selection for text classification via word embedding.
\newblock In {\em 2016 IEEE International Conference on Big Data Analysis (ICBDA)}, pages 1--5. IEEE, 2016.

\bibitem{sanh2019distilbert}
Victor Sanh, Lysandre Debut, Julien Chaumond, and Thomas Wolf.
\newblock Distilbert, a distilled version of bert: smaller, faster, cheaper and lighter.
\newblock {\em arXiv preprint arXiv:1910.01108}, 2019.

\bibitem{sheikhpour2017survey}
Razieh Sheikhpour, Mehdi~Agha Sarram, Sajjad Gharaghani, and Mohammad Ali~Zare Chahooki.
\newblock A survey on semi-supervised feature selection methods.
\newblock {\em Pattern recognition}, 64:141--158, 2017.

\bibitem{shu2017compressing}
Raphael Shu and Hideki Nakayama.
\newblock Compressing word embeddings via deep compositional code learning.
\newblock {\em arXiv preprint arXiv:1711.01068}, 2017.

\bibitem{tenenbaum2000global}
Joshua~B Tenenbaum, Vin~de Silva, and John~C Langford.
\newblock A global geometric framework for nonlinear dimensionality reduction.
\newblock {\em science}, 290(5500):2319--2323, 2000.

\bibitem{uysal2017sentiment}
Alper~Kursat Uysal and Yi~Lu Murphey.
\newblock Sentiment classification: Feature selection based approaches versus deep learning.
\newblock In {\em 2017 IEEE International Conference on Computer and Information Technology (CIT)}, pages 23--30. IEEE, 2017.

\bibitem{wang2019evaluating}
Bin Wang, Angela Wang, Fenxiao Chen, Yuncheng Wang, and C-C~Jay Kuo.
\newblock Evaluating word embedding models: Methods and experimental results.
\newblock {\em APSIPA transactions on signal and information processing}, 8:e19, 2019.

\bibitem{SIP-2023-0010}
Chengwei Wei, Yun-Cheng Wang, Bin Wang, and C.-C.~Jay Kuo.
\newblock An overview of language models: Recent developments and outlook.
\newblock {\em APSIPA Transactions on Signal and Information Processing}, 13(2), 2024.

\bibitem{wu2023brief}
Tianyu Wu, Shizhu He, Jingping Liu, Siqi Sun, Kang Liu, Qing-Long Han, and Yang Tang.
\newblock A brief overview of chatgpt: The history, status quo and potential future development.
\newblock {\em IEEE/CAA Journal of Automatica Sinica}, 10(5):1122--1136, 2023.

\bibitem{SIP-2023-0064}
Jintang Xue, Yun-Cheng Wang, Chengwei Wei, Xiaofeng Liu, Jonghye Woo, and C.-C.~Jay Kuo.
\newblock Bias and fairness in chatbots: An overview.
\newblock {\em APSIPA Transactions on Signal and Information Processing}, 13(2), 2024.

\bibitem{yang2022supervised}
Yijing Yang, Wei Wang, Hongyu Fu, and C-C~Jay Kuo.
\newblock On supervised feature selection from high dimensional feature spaces.
\newblock {\em APSIPA Transactions on Signal and Information Processing}, 11(1), 2022.

\bibitem{zhang2024evaluating}
Gaifan Zhang, Yi~Zhou, and Danushka Bollegala.
\newblock Evaluating unsupervised dimensionality reduction methods for pretrained sentence embeddings.
\newblock {\em arXiv preprint arXiv:2403.14001}, 2024.

\end{thebibliography}

\end{document}